\documentclass[conference,letterpaper,final]{IEEEtran}
\IEEEoverridecommandlockouts
\usepackage{cite}
\usepackage{graphicx}
\usepackage{textcomp}
\usepackage{xcolor}
\usepackage{algorithmic}
\usepackage[boxed]{algorithm2e}
\usepackage{amsmath,amssymb,amsfonts}
\usepackage{mathtools}

\usepackage{amsmath,amsfonts,bm}

\newcommand{\figleft}{{\em (Left)}}
\newcommand{\figcenter}{{\em (Center)}}
\newcommand{\figright}{{\em (Right)}}
\newcommand{\figtop}{{\em (Top)}}
\newcommand{\figbottom}{{\em (Bottom)}}

\def\eqref#1{equation~\ref{#1}}

\def\1{\bm{1}}

\DeclareMathAlphabet{\mathsfit}{\encodingdefault}{\sfdefault}{m}{sl}
\SetMathAlphabet{\mathsfit}{bold}{\encodingdefault}{\sfdefault}{bx}{n}

\def\sD{{\mathbb{D}}}

\def\sP{{\mathbb{P}}}

\def\sR{{\mathbb{R}}}

\def\sY{{\mathbb{Y}}}

\DeclareMathOperator*{\argmin}{arg\,min}

\usepackage{float}
\usepackage{subcaption}
\usepackage{forest,adjustbox}
\usepackage{tikz-qtree}
\usepackage[utf8]{inputenc} 
\usepackage[T1]{fontenc}    
\renewcommand{\baselinestretch}{1.1}
\usepackage{url}            
\usepackage{booktabs}       
\usepackage{multirow}
\usepackage{nicefrac}       
\usepackage{microtype}      

\newcommand\given[1][]{\:#1\vert\:}
\newcommand{\tabincell}[2]{\begin{tabular}{@{}#1@{}}#2\end{tabular}}

\makeatletter
\newcommand{\printfnsymbol}[1]{%
  \textsuperscript{\@fnsymbol{#1}}%
}
\makeatother

\def\BibTeX{{\rm B\kern-.05em{\sc i\kern-.025em b}\kern-.08em
    T\kern-.1667em\lower.7ex\hbox{E}\kern-.125emX}}

\begin{document}

\title{Neural Regression Trees}

\author{\IEEEauthorblockN{Shahan Ali Memon\IEEEauthorrefmark{2}, Wenbo Zhao\IEEEauthorrefmark{2} \thanks{\IEEEauthorrefmark{2}Equal Contribution. ISBN 978-1-7281-1985-4/19/\$31.00 \textcopyright2019 IEEE}, Bhiksha Raj and Rita Singh}
\IEEEauthorblockA{
Carnegie Mellon University\\
Pittsburgh, Pennsylvania 15213, United States\\
{\tt \{samemon, wzhao1, bhikshar, rsingh\}@andrew.cmu.edu}}
}

\maketitle

\begin{abstract}
Regression-via-Classification (RvC) is the process of converting a regression problem to a classification one. Current approaches for RvC use ad-hoc discretization strategies and are suboptimal. We propose a neural regression tree model for RvC. In this model, we employ a joint optimization framework where we learn optimal discretization thresholds while simultaneously optimizing the features for each node in the tree. We empirically show the validity of our model by testing it on two challenging regression tasks where we establish the state of the art.
\end{abstract}


\section{Introduction}
\label{sec:intro}

One of the most challenging problems in machine learning is that of  predicting a numeric attribute $y$ of a datum from other features $x$, a task commonly referred to as regression\footnote{Terminology that is borrowed from the statistics literature}.  
The relationship between the features and the predicted variable (which, for want of a better term, we will call a ``response'') is generally unknown and may not be deterministic. The general approach to the problem is to assume a formula for the relationship, and to estimate the details of the formula from training data.
Linear regression models assume a linear relationship between the features and the response. Other models such as neural networks assume a non-linear relationship. The problem here is that the model parameters that are appropriate for one regime of the data may not be appropriate for other regimes. Statistical fits of the model to the data will minimize a measure of the overall prediction error, but may not be truly appropriate for any specific subset of the data. Non-parametric regression models such as kernel regressions and regression trees attempt to deal with this by partitioning the feature space, and computing separate regressors within each partition. For high-dimensional data, however, any computed partition runs the risk of overfitting to the training data, necessitating simplifying strategies such as axis-aligned partitions~\cite{breiman2017classification,quinlan1986induction}.

An alternative strategy, and one that is explored in this paper, is to \emph{partition} the space based on the {\em response} variable. Formally, given a response variable $y$ that takes values in some range $(y_{\text{min}}, y_{\text{max}})$, we find a set of threshold values $t_0,\ldots,t_N$, and map the response variable into bins as $y \mapsto C_n~\text{if}~ t_{n-1} < y \le t_{n}$ for $n = 1, \ldots, N$. This process, which effectively converts the continuous-valued response variable $y$ into a categorical one $C$, is often referred to as {\em discretization}. The new response variable $C$ is, in fact, not merely categorical, but {\em ordinal}, since its values can be strictly ordered. In order to determine the value $y$ for any $x$ we must now find out which bin $C_n$ the feature $x$ belongs to; once the appropriate bin has been identified, the actual estimated $y$ can be computed in a variety of ways, e.g., the mean or median of the bin. Consequently, the problem of regression is transformed into one of classification. This process of converting a regression problem to a classification problem is uncommonly known as \textit{Regression-via-Classification (RvC)}~\cite{torgo1997regression}.

Naive implementation of RvC can, however, result in very poor regression. Inappropriate choice of bin boundaries $\{t_i\}$ can result in bins that are too difficult to classify (since classification accuracy actually depends on the distribution of feature $x$ within the bins). Alternatively, although permitting near-perfect classification, the bins may be too wide, and the corresponding ``regression'' may be meaningless. Ideally, the RvC method must explicitly optimize the bin boundaries for both classification accuracy {\em and} regression accuracy. In addition, the actual classifier employed cannot be ignored; since the decision boundaries are themselves variable, the classifier and the boundaries must conform to one another.

We propose a hierarchical tree-structured model for RvC, which addresses all of the problems mentioned above.
Jointly optimizing all the variables and classifiers involved against the classification accuracy and regression accuracy is a combinatorially hard problem. Instead of solving this directly, inspired by the original idea of regression trees \cite{quinlan1986induction}, we follow a greedy strategy of hierarchical binary partition of the response variable $y$, where each split is locally optimized. This results in a tree-structured RvC model with a classifier at each node. We employ a simple margin-based linear classifier for each classification; however, the {\em features} derived from the data may be optimized for classification. Moreover, the structure of our model affords us an additional optimization: instead of using a single generic feature for classification, we can now optimize the features extracted from the data {\em individually} for each classifier in the tree. Since we employ a neural network to optimize the features for classification, we refer to this framework as a \textit{Neural Regression Tree} (NRT).

To demonstrate the utility of the proposed approach we conduct experiments on a pair of notoriously challenging regression tasks: estimating the age and height of speakers from their voice. We show that our model performs significantly better than other regression models, including those that are known to achieve the current state-of-the-art in these problems.

\section{Related Work}
\label{sec:background}
\paragraph{Regression Trees}
Tree-structured models have been around for a long time. Among them, the most closely related are the regression trees. A regression tree is a regression function in which the partition is performed on features $x$ instead of response variable $y$.
The first regression tree algorithm was presented by \cite{morgan1963problems}, where they propose a greedy approach to fit a piece-wise constant function by recursively splitting the data into two subsets based on partition on the features $x$. The optimal split is a result of minimizing the impurity which defines the homogeneity of the split.
This algorithm sets the basis for a whole line of research on classification and regression trees. Improved algorithms include CART~\cite{breiman1984classification}, ID3~\cite{quinlan1986induction}, m5~\cite{quinlan1992learning}, and C4.5~\cite{salzberg1994c4}.
Recent work combines the tree-structure and neural nets to gain the power of both structure learning and representation learning. Such work includes the convolutional decision trees~\cite{laptev2014convolutional}, neural decision trees~\cite{xiao2017ndt,balestriero2017neural}, adaptive neural trees~\cite{tanno2018adaptive}, deep neural decision forests~\cite{kontschieder2015deep}, and deep regression forests~\cite{shen2017deep}.

We emphasize that there is a fundamental difference between our approach and the traditional regression tree based approaches: instead of making the split based on the feature space, our splitting criteria is based on the response variables, enabling the features to adapt to the partitions of response variables.

\paragraph{Regression via Classification (RvC)}
The idea for RvC was presented by~\cite{weiss1995rule}. Their algorithm is based on k-means clustering to categorize numerical variables.
Other conventional approaches~\cite{torgo1996regression,torgo1997regression} for discretization of continuous values are based on equally probable (EP) or equal width (EW) intervals, where EP creates a set of intervals with same number of elements, while EW divides into intervals of same range. These approaches are ad-hoc.
Instead, we propose a discretization strategy to learn the optimal thresholds by improving the neural classifiers.

\paragraph{Ordinal Regression} 
Because our model is essentially a method to discretize continuous values into ordered partitions, it can be somewhat compared to ordinal regression~\cite{mccullagh1980regression}. Ordinal regression is a class of regression analysis that operates on data where the response variable is categorical but exhibits an order relation. Naive approaches for ordinal regression often simplify the problem by ignoring the ordering information and treating the response variables as nominal categories. A slightly sophisticated method~\cite{gutierrez2016ordinal} uses decomposition of the response variable into several binary ones (such as via binary ordered partitions) and estimating them using multiple models. Another relevant class of approaches uses the threshold models~\cite{gutierrez2016ordinal}. These approaches resemble our approach in that they assume unobserved continuous response variables underlying the ordinal responses, and use thresholds to discretize them, where a variety of models (such as support vector machines and perceptrons) are used to model the underlying response variables.

Our proposed model shares many characteristics with these approaches.
However, one big difference is that in ordinal regression, the partitions are predefined by the domain problem and may not be optimized for statistical inference.
Our model, on the other hand, is based on a data-driven partition strategy where partitions are optimized for more discriminative representation of the data hierarchy and better performance at the inference time. 


\section{Neural regression tree}
\label{sec:method}

In this section, we formulate the neural regression tree model for optimal discretization of the response variables in RvC, and provide algorithms to optimize the model.
\subsection{Partition}
The key aspect of an RvC system is its method of partition $\Pi$. We define the partition $\Pi$ on a set $\sY \subset \sR$ as
\begin{align*}
\Pi(\sY) = \{C_1, \ldots, C_N\}
\end{align*}
satisfying $\bigcup_{n=1}^{N}C_n = \sY$ and $C_n$s are mutually disjoint. When acting on a $y \in \sY$, $\Pi(y) \coloneqq C_n \text{ subjected to } y \in C_n$.

\subsection{Model Formulation}
\label{ssec:mdl_form}
\begin{figure*}[!t]
  \begin{center}
  \scalebox{0.4}{
    \includegraphics{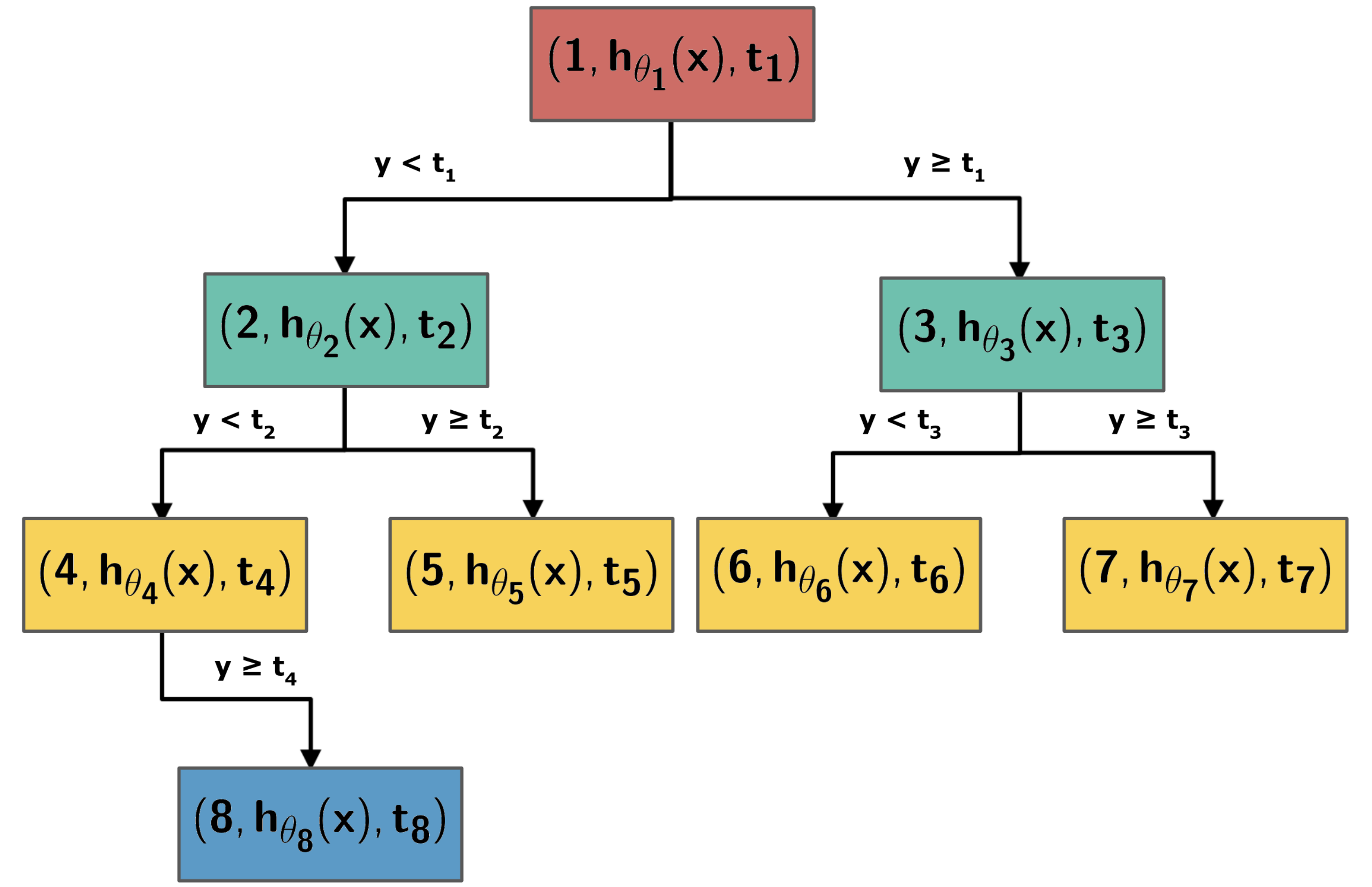}}
  \end{center}
  \caption{Illustration of neural regression tree. Each node is equipped with a neural classifier $h_{\theta}$. The splitting threshold $t$ is dependent on the response variable $y$ and is locally optimized.}
  \label{fig:binary_tree}
\end{figure*}

Formally, following the description of RvC in Section~\ref{sec:intro}, an RvC framework consists of two main rules: a classification rule and a regression rule. The classification rule classifies an input $x$ into disjoint bins, i.e., $h_{\theta}: x \mapsto \{C_1, \cdots, C_N\}$ with parameter $\theta$, where $C_n = \Pi(y)$ corresponds to $t_{n-1} < y \leq t_{n}$ for $n = 1, \ldots, N$. The regression rule $r: (x, C_n) \mapsto (t_{n-1}, t_n]$ maps the combination of input $x$ and class $C_n$ onto the interval $(t_{n-1}, t_n]$. Then, the combined RvC rule that predicts the value of the response variable for an input $x$ is
\begin{equation}
\hat{y}(x) = r(x, h_\theta(x)).
\label{eq:RvCregrquant}
\end{equation}


Instead of making a hard assignment of bins, alternatively, the classification rule $h_{\theta}$ may make a ``soft'' assignment by mapping an input $x$ onto the $N$-dimensional probability simplex, i.e., $h_{\theta}: x \mapsto \sP_N$ where $\sP_N$ represents the set of $N$-dimensional non-negative vectors whose entries sum to $1$.
The output of this classification rule is, therefore, the vector of \emph{a posteriori} probabilities over the $N$ classes, i.e., $h_{\theta}^n(x) = P(C_n \given x)$ where $h_{\theta}^n(x)$ represents the $n^{\text{th}}$ component of $h_{\theta}(x)$.
Hence, the RvC rule is given by
\begin{equation}
\hat{y}(x) = \mathbb{E}_{C_n}[r(x, C_n)] = \sum_{n=1}^{N} h_{\theta}^n(x) r_{C_n}(x),
\label{eq:RvCregr}
\end{equation}
where $r_{C_n}(\cdot) \coloneqq r(\cdot, C_n)$ fixes the second coordinate of $r$ at $C_n$.
Defining an error $\mathcal{E}(y, \hat{y}(x))$ between the true $y$ and the estimated $\hat{y}(x)$, our objective is to learn the thresholds $\{t_0, \ldots, t_N\}$ and the parameters $\{\theta_0, \ldots, \theta_N\}$ such that the expected error is minimized
\begin{align}
\{t_n^*\}, \{\theta_n^*\} \gets \argmin_{t, \theta} \mathbb{E}_x\left[\mathcal{E}(y, \hat{y}(x))\right].
\label{eq:min_error}
\end{align}
Note that the number of thresholds $(N + 1)$ too is a variable that may either be manually set or explicitly optimized. In practice, instead of minimizing the {\em expected} error, we minimize the {\em empirical} average error $\text{avg}(\mathcal{E}(y_i, \hat{y}(x_i)))$ computed over a training set.

However, joint optimization of $\{t_n\}$ and $\{\theta_n\}$ is a hard problem as it scales exponentially with $n$. To solve this problem we adopt the idea of divide and conquer and recast the RvC problem in terms of a binary classification tree, where each of the nodes in the tree is greedily optimized.
The structure of the proposed binary tree is shown in Figure~\ref{fig:binary_tree}. 

We now describe the tree-growing algorithm. For notational convenience the nodes have been numbered such that for any two nodes $n_1$ and $n_2$, if $n_1 < n_2$, $n_1$ occurs either to the left of $n_2$ or above it in the tree.
Each node $n$ in the tree has an associated threshold $t_n$, which is used to partition the data into its two children $n'$ and $n''$ (we will assume w.l.o.g. that $n' < n''$).  A datum $(x,y)$ is assigned to the ``left'' child $n'$ if $y < t_n$, and to the ``right'' child $n''$ otherwise. The actual partitions of the response variable are the leaves of the tree.
To partition the data, each node carries a classifier $h_{\theta_n}: x \mapsto \{n', n''\}$, which assigns any instance with features $x$ to one of $n'$ or $n''$. 
In our instantiation of this model, the classifier $h_{\theta_n}$ is a neural classifier that not only classifies the features but also adapts and refines the features to each node.

Given an entire tree along with all of its parameters and an input $x$, we can compute the {\em a posteriori} probability of the partitions (i.e. the leaves) as follows. For any leaf $l$, let $l_0 \rightarrow \cdots \rightarrow l_p$ represent the chain of nodes from root $l_0$ to the leaf itself $l_p \equiv l$. The {\em a posteriori} probability of the leaf is given by $P(l \given x) = \prod_{r=1}^p P(l_r \given l_{r-1}, x)$, where each $P(l_r \given l_{r-1}, x)$ is given by the neural classifier on node $l_{r-1}$. Substitution into (\ref{eq:RvCregr}) yields the final regressed value of the response variable
\begin{equation}
\hat{y}(x) = \sum_{l \in \text{leaves}} P(l \given x) r_l(x),
\label{eq:RvCregrtree}
\end{equation}
where $r_l(x)\coloneqq r(x, l)$, in our setting, is simply the mean value of the leaf bin. Other options include the center of gravity of the leaf bin, using a specific regression function, etc.

\subsection{Learning the Tree}
\label{ssec:learn_tree}
We learn the tree in a greedy manner, optimizing each node individually. The procedure to optimize an individual node $n$ is as follows. Let $\sD_n = \{(x_i, y_i)\}$ represent the set of training instances arriving at node $n$. Let $n'$ and $n''$ be the children induced through threshold $t_n$. In principle, to locally optimize $n$, we must minimize the average regression error ${\mathcal E}(\sD_n; t_n, \theta_n) = \text{avg}\left({\mathcal E}(y, \hat{y}_n(x))\right)$ between the true response values $y$ and the estimated response $\hat{y}_n(x)$ computed using only the subtree with its root at $n$. In practice, ${\mathcal E}(\sD_n; t_n, \theta_n)$ is not computable, since the subtree at $n$ is as yet unknown. Instead, we will approximate it through the {\em classification} accuracy of the classifier at $n$, with safeguards to ensure that the resultant classification is not trivial and permits useful regression. 

Let $y(t_n) = \text{sign}(y - t_n)$ be a binary indicator function that indicates if an instance $(x, y)$ has to be assigned to child $n'$ or $n''$. Let ${\mathcal E}(y(t_n), h_{\theta_n}(x))$ be a qualifier of the classification error (which can be binary cross entropy loss, hinge loss, etc.) for any instance $(x, y)$.
We define the classification loss at node $n$ as
\begin{equation}
E_{\theta_n, t_n} = \frac{1}{|\sD_n|} \sum_{(x, y) \in \sD_n}{\mathcal E}(y(t_n), h_{\theta_n}(x)),
\label{eq:closs}
\end{equation}
where $|\sD_n|$ is the size of $\sD_n$. The classification loss (\ref{eq:closs}) cannot be directly minimized w.r.t $t_n$, since this can lead to trivial solutions, e.g., setting $t_n$ to an extreme value such that all data are assigned to a single class. While such a setting would result in perfect classification, it would contribute little to the regression. To prevent such solutions, we include a {\em triviality} penalty $\mathcal{T}$ that attempts to ensure that the tree remains balanced in terms of the number of instances at each node. For our purpose, we define the triviality penalty at any node as the entropy of the distribution of instances over the partition induced by $t_n$ (other triviality penalties such as the Gini index~\cite{breiman1984classification} may also apply though)
\begin{equation}
\mathcal{T}(t_n) = -p(t_n)\log p(t_n) - (1-p(t_n))\log(1 - p(t_n)),
\label{eq:trivialitiy}
\end{equation}
where 
\[
p(t_n) = \frac{\sum_{(x,y)\in \sD_n} (1 + y(t_n))}{2|\sD_n|}.
\label{eq:entropy}
\]
The overall optimization of node $n$ is performed as
\begin{align}
\theta_n^*, t_n^*= \argmin_{\theta_n, t_n}\lambda E_{\theta_n, t_n}  + (1 - \lambda) \mathcal{T}(t_n),
\label{eq:obj}
\end{align}
where $\lambda \in (0, 1)$ is used to assign the relative importance of the two components of the loss, and is a hyper-parameter to be tuned.

In the optimization of (\ref{eq:obj}), the loss function depends on $t_n$ through $y(t_n)$, which is a discontinuous function of $t_n$. To get around this difficulty of optimizing (\ref{eq:obj}), we have two possible ways: the {\bf scan method} and the {\bf gradient method}. In the first, we can {\em scan} through all possible values of $t_n$ to select the one that results in the minimal loss.
Alternatively, a faster gradient-descent approach is obtained by making the objective differentiable w.r.t. $t_n$. Here the discontinuous function $\text{sign}(y - t_n)$ is approximated by a differentiable relaxation: $y(t_n) = 0.5(\tanh(\beta(y-t_n))+1)$, where $\beta$ controls the steepness of the function and must typically be set to a large value ($\beta=10$ in our settings) for close approximation.  The triviality penalty is also redefined (to be differentiable w.r.t. $t_n$) as the proximity to the median $\mathcal{T}(t_n) = \|t_n - \text{median}(y \given (x,y) \in \sD_n)\|_2$, since the median is the minimizer of (\ref{eq:trivialitiy}). We use coordinate descent to optimize the resultant loss.

Once optimized, the data $\sD_n$ at $n$ are partitioned into $n'$ and $n''$ according to the threshold $t^*_n$, and the process proceeds recursively down the tree. The growth of the tree may be continued until the regression performance on a held-out set saturates.
Algorithm~\ref{alg:nrt} describes the entire training algorithm.
\SetKwInOut{Parameter}{Parameter}
\SetKwProg{Fn}{Function}{}{}
\SetKwFunction{BuildTree}{BuildTree}
\SetKwFunction{Partition}{Partition}
\SetKwFunction{FeatureExtractor}{FeatureExtractor}
\SetKwFunction{NeuralClassifier}{NeuralClassifier}
\begin{algorithm}[!t]
	\KwIn{$\sD$}
	\Parameter{$\{t_n\}, \{\theta_n\}$}
	\KwOut{$\{t_n^*\}, \{\theta_n^*\}$}
	
    \Fn{\BuildTree{$\sD_n$}}{
        Initialize $t_n$, $\theta_n$
        
   		$t_n^*, \theta_n^* \gets$ \NeuralClassifier{$\sD_n, t_n, \theta_n$}
   		
        $\sD_{n'}, \sD_{n''} \gets$ \Partition{$\sD_n, t_n^*$}
        
   		\For{$\sD_n$ in $\{\sD_{n'}, \sD_{n''}\}$}{
           	\eIf{$\sD_{n}$ is pure}{
           	    continue
           	}{
   				\BuildTree{$\sD_n$}
            }
   		}
 	}

	\BuildTree{$\sD$}
\caption{{\bf Learning neural regression tree.} The tree is built recursively. For each node $n$, it adapts and classifies the features, and partitions the data based on the locally optimal classification threshold.} \label{alg:nrt}
\end{algorithm}

\section{Experiments}
\label{sec:experiment}
We consider two regression tasks in the domain of speaker profiling---age estimation and height estimation from voice. The two tasks are generally considered two of the  challenging tasks in the speech literature~\cite{kim2007age,metze2007comparison,li2010combining,dobry2011supervector,lee2012performance,bahari2014speaker,barkana2015new,poorjam2015height,fu2008human}. 

We compare our model with 1) a regression baseline using the support vector regression (SVR)~\cite{basak2007support}, 2) a regression tree baseline using classification and regression tree (CART)~\cite{breiman1984classification}, and 3) a neural net baseline with multilayer perceptron (MLP) structure. Furthermore, in order to show the effectiveness of the ``neural part'' of our NRT model, we further compare our neural regression tree with a third baseline 4) regression tree with the support vector machine (SVM-RT).

\subsection{Data}
\label{ssec:expr_setup}
\begin{table}[!t]
	\caption{Fisher Dataset Partitions}
	\label{tab:fisher_stat}
	\centering
	\begin{tabular}{lrr}
		\toprule
		& \multicolumn{2}{c}{\textbf{\# of Speakers / Utterances}} \\  
        & \multicolumn{1}{c}{Male} & \multicolumn{1}{c}{Female} \\
		\midrule
        Train & 3,100 / 28,178 & 4,813 / 45,041 \\
        Dev & 1,000 / 9,860 & 1,000 / 9,587 \\
        Test & 1,000 / 9,813 & 1,000 / 9,799 \\
		\bottomrule
	\end{tabular}
\end{table}

\begin{figure}[!t]
  \centering
  \includegraphics[width=1.0\linewidth]{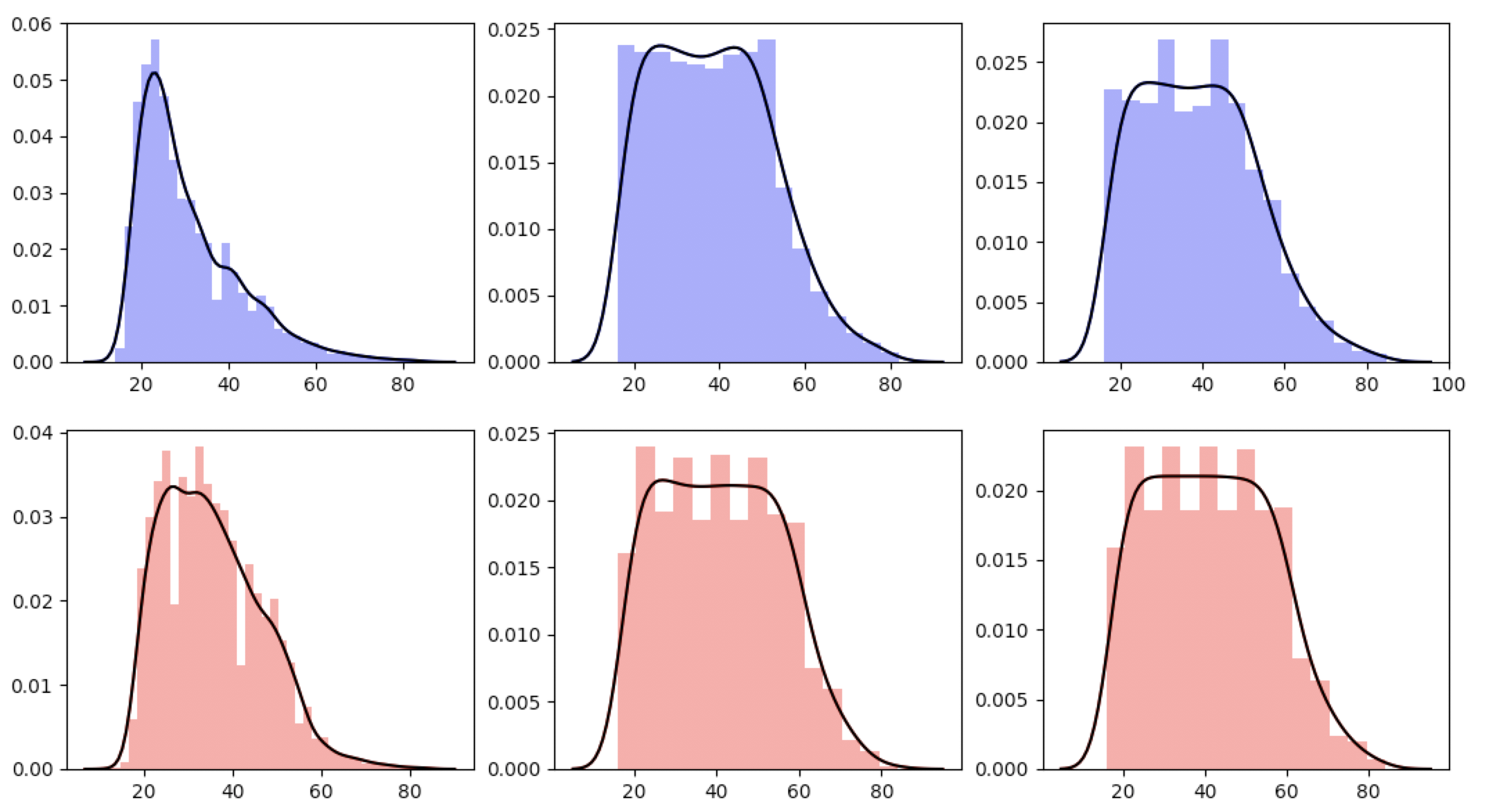}
  \caption{Age distribution (in percentages) for male~\figtop~and female~\figbottom~speakers for the Fisher database for train~\figleft, development~\figcenter~and test~\figright~sets. The horizontal axis is age.}
  \label{fig:fisher-dist}
\end{figure}

\begin{table}[!t]
	\caption{NIST-SRE8 Dataset Stats}
	\label{tab:sre8_stat}
	\centering
	\begin{tabular}{lrr}
		\toprule
		& \multicolumn{2}{c}{\textbf{\# of Speakers / Utterances}} \\  
        & \multicolumn{1}{c}{Male} & \multicolumn{1}{c}{Female} \\
		\midrule
         & 384 / 33,493 & 651 / 59,530\\
		\bottomrule
	\end{tabular}
\end{table}

\begin{figure}[!t]
\centering
  \begin{subfigure}[b]{0.45\linewidth}
    \includegraphics[width=\textwidth]{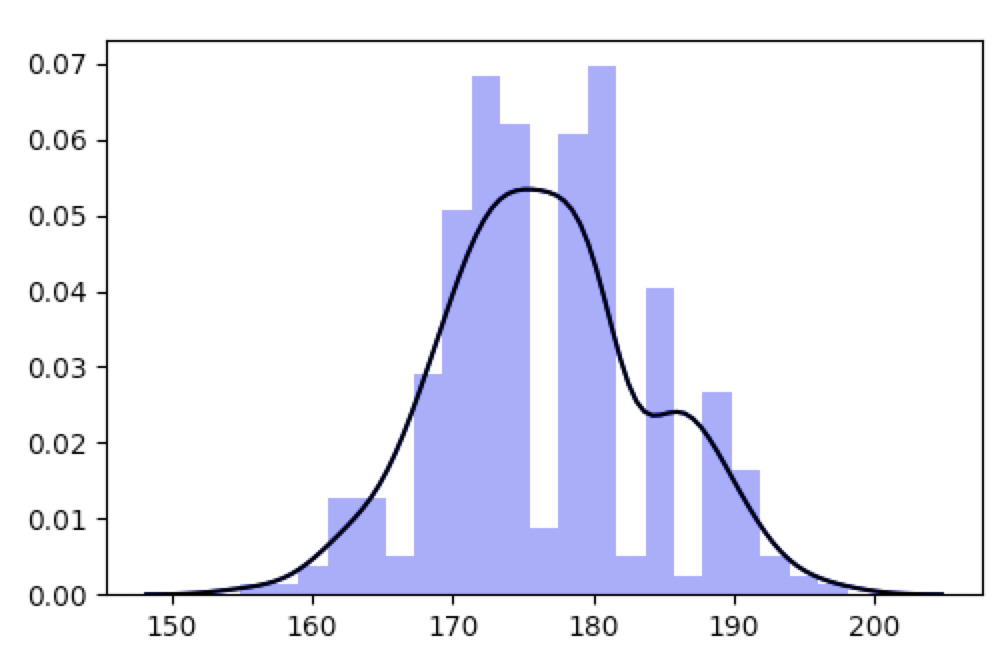}
    \label{fig:1}
  \end{subfigure}
  \begin{subfigure}[b]{0.45\linewidth}
    \includegraphics[width=\textwidth]{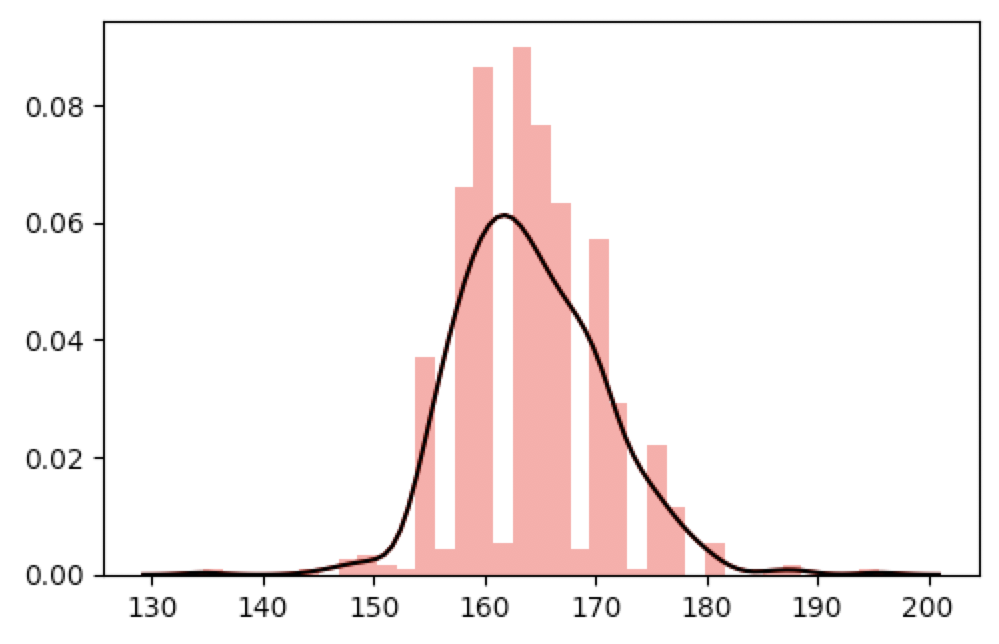}
    \label{fig:2}
  \end{subfigure}
  \caption{Height distribution (in percentages) for male~\figleft~and female~\figright~speakers for the NIST-SRE8 database. The horizontal axis is height.}   \label{fig:sre8-stat}
\end{figure}

To promote a fair comparison, we select two well-established public datasets in the speech community. For age estimation, we use the Fisher English corpus~\cite{cieri2004fisher}. It consists of a 2-channel conversational telephone speech for $11,971$ speakers, comprising of a total of $23,283$ recordings. After removing $58$ speakers with no age specified, we are left with $11,913$ speakers with $5,100$ male and $4,813$ female speakers. To the best of our knowledge, the Fisher corpus is the largest English language database that includes the speaker age information to be used for the age estimation task.
The division of the data for the age estimation task is shown in Table~\ref{tab:fisher_stat}. The division is made through stratified sampling such that there is no overlap of speakers and all age groups are represented across the splits. Furthermore, Figure~\ref{fig:fisher-dist} shows the age distribution of the database for the three splits (train, development, test) in relation to the Table~\ref{tab:fisher_stat}. 

For height estimation, we use the NIST speaker recognition evaluation (SRE) 2008 corpus~\cite{kajarekar2009sri}. We only obtain heights for $384$ male speakers and $651$ female speakers from it. Because of such data scarcity issues, we evaluate this task using cross-validation. Table~\ref{tab:sre8_stat} and Figure~\ref{fig:sre8-stat} show the statistics for the NIST-SRE8 database.

Since the recordings for both datasets have plenty of silences and the silences do not contribute to the information gain, Gaussian based voice activity detection (VAD) is performed on the recordings. Then, the resulting recordings are segmented to one-minute segments.

To properly represent the speech signals, we adopt one of the most effective and well-studied representations, the i-vectors~\cite{dehak2011front}.
I-vectors are statistical low-dimensional representations over the distributions of spectral features, and are commonly used in state-of-the-art speaker recognition systems~\cite{sadjadi2016speaker} and age estimation systems~\cite{shivakuma14simplifiedivector,grzybowska2016speaker}.
Respectively, $400$-dimensional and $600$-dimensional i-vectors are extracted for Fisher and SRE datasets using the state-of-the-art speaker identification system~\cite{dhamyal2018optim}.

\begin{table}[!t]
	\caption{Model Specifications}
	\label{tab:mdl_spec}
	\centering
	\begin{tabular}{lll}
		\toprule
        {\bf Model} & \multicolumn{2}{c}{{\bf Specification}} \\
        & \multicolumn{1}{c}{Age} & \multicolumn{1}{c}{Height} \\
		\midrule
        NRT & \tabincell{l}{Linear: (400, 1000) \\ Linear: (1000, 1000) \\ Linear: (1000, 1) \\ Nonlin.: ReLU \\ Optim.: Adam (lr 0.001)} & \tabincell{l}{(Same as Age with \\ input dim. of 600)} \\
        \midrule
        SVM-RT & \tabincell{l}{Kernel: RBF \\ Regul.: $\ell_1$\\
        Optim.: Scan (Sec.~\ref{ssec:learn_tree})} & \tabincell{l}{(Same as Age with \\ linear kernel)} \\
        \midrule
        SVR & \tabincell{l}{Kernel: RBF \\ Regul.: $\ell_1$ \\ } & \tabincell{l}{Kernel: Linear \\ Regul.: $\ell_1$ \\} \\
        \midrule
        CART & \tabincell{l}{Criteri.: MSE} & \tabincell{l}{Criteri.: MSE}\\
        \midrule
        MLP & \tabincell{l}{Linear: (400, 512) \\ Linear: (512, 1) \\ Nonlin.: ReLU \\ Optim.: Adam (lr 0.01)} & \tabincell{l}{Linear: (600, 2048) \\ Linear: (2048, 1) \\ Nonlin.: ReLU \\ Optim.: Adam (lr 0.005)} \\
		\bottomrule
	\end{tabular}
\end{table}

\subsection{Model}

\begin{table*}[!t]
\centering
\setlength\tabcolsep{15pt}
\caption{Performance evaluation of neural regression tree and baselines.}
\label{tab:results}
\begin{tabular}{cclrrrrr}
\toprule 
	Task & Dataset & \textbf{Methods} & \multicolumn{2}{c}{\textbf{Male}} && \multicolumn{2}{c} {\textbf{Female}}\\
\cmidrule{4-5} \cmidrule{7-8}
    & & & MAE & RMSE && MAE & RMSE \\
\midrule
\multirow{4}{*}{Age} & \multirow{4}{*}{Fisher} & SVR & 9.22 & 12.03 && 8.75 & 11.35 \\
    & & CART & 11.73 & 15.22 && 10.75 & 13.97 \\
    & & MLP  &  9.06 & 11.91 &&  8.21 & 10.75 \\
    \cmidrule{3-8}
	& & SVM-RT & 8.83 & 11.47 && 8.61 & 11.17 \\
	& & NRT & {\bf 7.20} & {\bf 9.02} && {\bf 6.81} & {\bf 8.53} \\
\midrule
\multirow{4}{*}{Height} & \multirow{4}{*}{SRE} & SVR & 6.27 & 6.98 && 5.24 & {\bf 5.77} \\
    & & CART & 8.01 & 9.34 && 7.08 & 8.46 \\
    & & MLP  & 8.17 &10.92 && 7.46 & 9.47 \\
    \cmidrule{3-8}
	& & SVM-RT & 5.70 & 7.07 && 4.85 & 6.22 \\
	& & NRT & {\bf 5.43} & {\bf 6.40} && {\bf 4.27} & 6.07 \\
\bottomrule
\end{tabular}
\end{table*}

The specifications for our model and the baseline models are shown in Table~{\ref{tab:mdl_spec}}.
The proposed NRT is a binary tree with neural classification models as explained in Section~\ref{ssec:mdl_form}, where the neural classifiers are $3$-layer ReLU neural networks.
Each model is associated with a set of hyper-parameters (e.g., the $\lambda$ in (\ref{eq:obj}), the number of neurons and layers for the neural nets, the margin penalty and kernel bandwidth for SVM and SVR, the depth for CART, etc.) that have to be tuned on the development set. These hyper-parameters control the complexity and generalization ability of the corresponding models. We tune them based on the bias-variance trade-off until the best performance on development set has been achieved.

\subsection{Results}

To measure the performance of our models on the age and height estimation tasks, we use the mean absolute error (MAE) and the root mean squared error (RMSE) as evaluation metrics. The results are summarized in Table~\ref{tab:results}. To reduce the effect of weights initialization on the performance of models consisting neural nets, we run those models multiple (10) times with different initialization, and report the average performance error.

For both age and height estimation, we observe that the proposed neural regression tree model generally outperforms other baselines in both MAE and RMSE, except that for height task, the neural regression tree has slightly higher RMSE than SVR, indicating higher variance. This is reasonable as our NRT does not directly optimize on the mean squared error. Bagging or forest mechanisms may be used to reduce the variance.
Furthermore, with the neural classifier in NRT being replaced by an SVM classifier (SVM-RT), we obtain higher errors than NRT, demonstrating the effectiveness of the neural part of the NRT as it enables the features to refine with each partition and adapt to each node.
Nevertheless, SVM-RT still yields smaller MAE and RMSE values than SVR and CART, and is on par with the MLP on the age task; on the height task, SVM-RT outperforms SVR, CART and MLP in terms of MAE values while also showing relatively small variances. This consolidates our claim that even without the use of a neural network, our model can find optimal thresholds for the discretization of response variables.
On the other hand, this also confirms that using neural nets without the tree adaptation only contributes to a small portion of the performance gain provided that the neural nets generalize well.
Additionally, we observe that a simple-structured MLP, as compared to the MLP component in NRT, is required to obtain reasonable performance---a more complex-structured MLP would not generalize well to the test set and yield high estimation bias. This, in turn, implies that our NRT model can employ high-complexity neural nets to adapt the features to be more discriminative as the discretization refines, while at the same time maintain the generalization ability of the model.

To test the significance of the results, we further conduct pairwise statistical significance tests. We hypothesize that the errors achieved from our NRT method are significantly smaller than the closest competitor SVR.
Paired t-test for SVR \emph{v.s.} SVM-RT and SVM-RT \emph{v.s.} NRT yield p-values less than $2.2 \times 10^{-16}$, indicating significant improvement. Similar results are obtained for height experiments as well. Hence, we validate the significance of the performance improvement of our NRT method on estimating ages and heights over the baseline methods.

\subsection{Node-based Error Analysis}

The hierarchical nature of our formulation allows us to analyze our model on every level and every node of the tree in terms of its classification and regression error.
Figure~\ref{fig:mae-tree-result} shows the per-level regression errors in terms of MAE for female and male speakers, where the nodes represent the age thresholds used as splitting criteria at each level, and the edges represent the regression errors.
We notice that regression error increases from left to right for both female and male speakers (except the leftmost nodes where the behavior does not follow possibly due to data scarcity issues), meaning the regression error for the younger speakers is lower than the error for older speakers.
In other words, our model is able to discriminate better between younger speakers. This is in agreement with the fact that the vocal characteristics of humans undergo noticeable changes during earlier ages, and then relatively stabilize for a certain age interval~\cite{stathopoulos2011changes}.
Hence, the inherent structural properties of our model not only improve the overall regression performance as we see in the previous section, but in the case of age estimation, also model the real world phenomenon.
\begin{figure}[!t]
\centering
{\footnotesize
\scalebox{1.0}{
\begin{forest}
for tree={circle,l sep=15pt,minimum width=0.8cm}
[43, red, draw  
    [31, red, edge label={node[midway,left] {5.4}}, draw
      [24,red, edge label={node[midway,left] {5.79}}, draw
      	[,edge label={node[midway,left] {6.9}}, edge={dashed}]
        [,edge label={node[midway,right] {4.9}}, edge={dashed}]
      ]
      [37, red, edge label={node[midway,right] {5.14}}, draw
      	[,edge label={node[midway,left] {5.3}}, edge={dashed}]
        [,edge label={node[midway,right] {4.96}}, edge={dashed}]
      ]
    ]
    [55, red, edge label={node[midway,right] {8.9}}, draw
      [49, red, edge label={node[midway,left] {5.83}}, draw
      	[,edge label={node[midway,left] {5.02}}, edge={dashed}]
        [,edge label={node[midway,right] {6.63}}, edge={dashed}]
      ] 
      [61, red, edge label={node[midway,right] {12.23}}, draw
      	[,edge label={node[midway,left] {8.97}}, edge={dashed}]
        [,edge label={node[midway,right] {15.64}}, edge={dashed}]
      ]
  	] 
]
\end{forest}
}
}
{\footnotesize
\scalebox{1.0}{
\begin{forest}
for tree={circle, l sep=15pt,minimum width=0.8cm}
[40, blue, draw  
    [29, blue, edge label={node[midway,left] {5.57}},draw
      [23, blue, edge label={node[midway,left] {5.91}},draw 
      	[,edge label={node[midway,left] {7.25}}, edge={dashed}]
        [,edge label={node[midway,right] {4.81}}, edge={dashed}]
      ] 
      [35, blue,edge label={node[midway,right] {5.24}}, draw
      	[,edge label={node[midway,left] {4.89}}, edge={dashed}]
        [,edge label={node[midway,right] {5.67}}, edge={dashed}]
      ]
    ]
    [50, blue, edge label={node[midway,right] {9.77}}, draw
      [45, blue, edge label={node[midway,left] {6.92}}, draw
      	[,edge label={node[midway,left] {6.51}}, edge={dashed} ]
        [,edge label={node[midway,right] {7.32}}, edge={dashed}]
      ] 
      [56, blue, edge label={node[midway,right] {12.44}}, draw
      	[,edge label={node[midway,left] {8.9}}, edge={dashed}]
        [,edge label={node[midway,right] {15.61}}, edge={dashed}]
      ] 
  ] 
]
\end{forest}
}
}
\caption{Regression errors for different age groups for female~\figleft~and male~\figright~for the task of speaker age estimation. Each node represents the age threshold used as splitting criterion, and each edge represents the regression error in terms of MAE.} \label{fig:mae-tree-result}
\end{figure}

\subsection{Limitations}
\label{ssec:limitations}
We acknowledge that our model might not be ubiquitous in its utility across all regression tasks. Our hypothesis is that it works well with tasks that can benefit from a partition based formulation. We empirically show that to be true for two such tasks above. However, in future we would like to test our model for other standard regression tasks.
Furthermore, because our model formulation inherits its properties from the regression-via-classification (RvC) framework, the objective function is optimized to reduce the \emph{classification error} rather than the \emph{regression error}. This limits us in our ability to directly compare our model to other regression methods. In future, we intend to explore ways to directly minimize the regression error while employing the RvC framework.

\section{Conclusions}
\label{sec:concl}
In this paper, we proposed Neural Regression Trees (NRT) for optimal discretization of response variables in regression-via-classification (RvC) tasks. It targeted the two challenges in traditional RvC approaches: finding optimal discretization thresholds, and selecting optimal set of features. We developed a discretization strategy by recursive binary partition based on the optimality of neural classifiers.
Furthermore, for each partition node on the tree, it was able to locally optimize features to be more discriminative.
We proposed a scan method and a gradient method to optimize the tree. The proposed NRT model outperformed baselines in age and height estimation experiments, and demonstrated significant improvements. 

%
%

\bibliographystyle{IEEEtran}
\bibliography{ijcnn_2019}
\end{document}